\documentclass[runningheads]{llncs}
\usepackage{graphicx}
\usepackage{amsmath,amssymb} 
\usepackage{color}
\usepackage[colorlinks,
            linkcolor=black,
            anchorcolor=black,
            citecolor=black]{hyperref}
\usepackage{multirow}
\usepackage{makecell}
\usepackage{tablefootnote}
\usepackage{enumerate}
\usepackage{xcolor}

\newcommand{\NOTE}[1]{\textcolor{red}{[NOTE: #1]}}

\newcommand{\comments}[1]{}

\newcommand{\wqz}[1]{\textcolor{black}{#1}}
\newcommand{\REPLY}[1]{\textcolor{cyan}{[REPLY: #1]}}
\newcommand{\eg}{\textit{e.g.}}
\newcommand{\Eg}{\textit{E.g.}}

\begin{document}

\title{Gated Hierarchical Attention for Image Captioning} 
\titlerunning{Gated Hierarchical Attention for Image Captioning} 


\author{Qingzhong Wang \and Antoni B. Chan}
%

\authorrunning{Q. Wang \and A. B. Chan} 


\institute{Department of Computer Science, City University of Hong Kong \email{qingzwang2-c@my.cityu.edu.hk} \quad \email{abchan@cityu.edu.hk}}

\maketitle

\begin{abstract}
Attention modules connecting encoder and decoders have been widely applied in the field of object recognition, image captioning, visual question answering and neural machine translation, and significantly improves the performance. In this paper, we propose a bottom-up gated hierarchical attention (GHA) mechanism for image captioning. Our proposed model employs a CNN as the decoder which is able to learn different concepts at different layers, and apparently, different concepts correspond to different areas of an image. Therefore, we develop the GHA in which low-level concepts are merged into high-level concepts and simultaneously low-level attended features pass to the top to make predictions. Our GHA significantly improves the performance of the model \wqz{that} only applies one level attention, \eg, the CIDEr score increases from 0.923 to 0.999, which is comparable to the state-of-the-art models that employ attributes boosting and reinforcement learning (RL). We also conduct extensive experiments to analyze the CNN decoder and our proposed GHA, and we find that deeper decoders cannot obtain better performance, and when the convolutional decoder becomes deeper the model is likely to collapse during training. Code is available: \url{https://github.com/qingzwang/GHA-ImageCaptioning}.

\keywords{Hierarchical Attention \and Image Captioning \and Convolutional Decoder.}
\end{abstract}
\section{Introduction}\label{intro}

Image captioning aims to automatically generate sentences that are able to describe images. To achieve this goal, an image captioning model should contain at least three parts: 1) a vision module, which extracts features from images, 2) a language module, which is used to model the sentences, 3) a connection module, which is applied to fuse the vision and language context information. 

Recently, CNNs are the most popular vision module, such as VGG nets \cite{vggnet}, Google nets \cite{inceptionv3} and residual nets \cite{resnet} (in this paper, we call them Image-CNNs). It is believed that introducing more information benefits the performance, and hence some models employ object detection 
or transfer image features into attributes to obtain more details or semantic information of an image \cite{bottomup,visconcept,semanticatt,attributes,attributesboosting,SCN}. However, applying object detection or attributes boosting methods requires more annotations, such as the bounding boxes of the objects and their attributes, categories and actions, which is difficult to obtain. For the language module, LSTMs dominate the field of sentence generation \cite{NIC,spatt,philstm,hierlstm}. 
Almost all previous image captioning models employ LSTMs or their variants. LSTMs apply gates and a memory cell to fuse the information of a sequence of words. In contrast, treating the sequence of word embeddings as a matrix, CNNs stack multiple convolution layers to merge words together \cite{glulanguage} into higher-level concepts (in this paper, we call these Word-CNNs). While it is difficult to implement LSTMs in a parallel way, because there is a recurrent path during the computation, Word-CNNs only stack several layers and each layer is computed in parallel, increasing the speed. For example, to obtain the representation of 12 words, an LSTM needs 12 steps, while a Word-CNN with kernel size 3 only needs 6 layers. In terms of the connection module, a naive approach is to directly concatenate the image features and language context \cite{mrnn}. However different words correspond to different areas of an image, which limits the accuracy of the naive approach.
The NIC model \cite{NIC} first transfers the image features into another space at the first LSTM step and it outperforms m-RNN model \cite{mrnn}. However, \cite{NIC,mrnn} are not able to learn the correspondence between words and image areas. To address this issue, \cite{spatt} introduces an attention mechanism to the LSTM-based captioning models, which is able to learn the relationship between words and image areas, and the performance improves. 

The connection module plays a crucial role in image captioning models and the attention mechanism leads to both better performance and also better interpretation. For example, when a captioning model generates the word ``person'', the model will pay more attention to the area of the person -- moreover it can also learn to find the appropriate image areas corresponding to the words that do not refer to objects.  
Word-CNNs are able to merge words to form high-level concepts, which mimics the tree structure of the sentence. Here we assume that each concept should correspond to an area of an image (e.g., as noted by image-text retrieval models \cite{mcnn,relstm}), 
and that multi-level attention is able to provide more accurate and detailed information \cite{recurrentatt,learn2payatt,vqaatt,dualrecurrentatt}. 

Inspired by \cite{mcnn,relstm}, we propose a gated hierarchical attention mechanism (GHA) for image captioning. The difference between our proposed GHA and other popular attention mechanisms is that GHA allows multi-level interaction between image encoder and language decoder, and the gating mechanism is applied to select low-level attention features and pass them to the high-level ones.
%
%
Applying the proposed GHA, we obtain an increase of CIDEr score by 8.2\% and SPICE by 8.6\% 
on MSCOCO dataset, and obtain comparable results to the state-of-the-art models. Another contribution of this paper is that we conduct extensive experiments to analyze the proposed GHA and Word-CNNs.

\section{Related Work}\label{rw}

\subsection{Image Captioning with Attention Mechanisms}

Much work has shown that attention is able to significantly improve the captioning results \cite{spatt,semanticatt,areasatt}. In a captioning model, attention plays the role of connecting the image encoder and language decoder. The attention module normally takes the image feature map and context as input, and outputs a feature vector. In \cite{spatt}, an MLP is applied to learn the correspondence between words and image areas. When generating the current word, the attention module first computes an image representation by using the previous hidden state of the LSTM, and then the image representation and current hidden state are used to make predictions. After training, the attention module is able to show interpretable results. However, this attention mechanism computes attention feature for each word which is unreasonable because some words in the captions do not correspond to any object in an image.  For example, the words ``a'' and ``of'' do not refer to any object, and therefore they are more likely to depend on the context instead of the image. To solve this problem, \cite{when2look} introduces a visual sentinel to the attention module, which is a gate that decides whether the context or the attention feature is used to predict the current word. Other attention variants apply semantic, attribute and object detection results to introduce more information \cite{semanticatt,bottomup,areasatt}. Generally, using more information results in better performance. \wqz{In \cite{AAAIpaper}, multiple convolutional maps are employed to improve video captioning, which is similar to our GHA model. The decoder in \cite{AAAIpaper} is an LSTM and to compute the attention maps at each time step, the same LSTM hidden state is applied. In contrast, our GHA model employs convolutional decoders, and therefore different concept representations are applied to compute attention maps, which reveals more about the relationships between concepts and image regions.}

Moreover, our GHA model is able to obtain more visual information and hierarchically pass it from the bottom to the top level, while other models only employ attention at one layer. Although some models use gates, such as \cite{when2look}, the goals are different. The gates in our GHA filter out low-level information that cannot benefit prediction of the current word. In contrast, in \cite{when2look}, the gates decide whether the current word should depend on the image feature or the previous words. 
Moreover, in \cite{when2look}, all channels share the same gate, whereas, in our proposed GHA, a separate gate is used for each channel.

\subsection{Word-CNNs for NLP}

Word-CNNs have been widely applied in the field of NLP (e.g., text classification \cite{cnntextclass,verydeeptextclass}, language modeling \cite{glulanguage}, and machine translation \cite{convseq2seq}) due to several advantages:
1) convolution is faster than LSTMs, because convolution can be easily computed in parallel;
2) CNNs imitate a tree structure of the sentence, which is able to benefit solving NLP tasks. 

In \cite{languagecnn}, a language CNN is applied to improve the long-term memory of LSTMs, however it also employs an LSTM to generate captions. \cite{convimagecap,cnnpluscnn} drop LSTMs and only apply CNNs to generate captions, which shows faster training. Although \cite{convimagecap,cnnpluscnn} both compute attention features at each convolutional layer, they use simple attention mechanisms, which just concatenates or adds low-level attended visual features to the high-level concept features. One disadvantage of this simple approach is that when the model becomes deeper, the influence of low-level features could vanish. Another disadvantage is that sometimes the predictions only depend on high-level features, and introducing low-level features could result in unexpected predictions. In contrast, our GHA employs a gated recurrent unit (GRU) to learn to memorize or forget low-level features. Our proposed GHA first fuses the low-level concept features and the corresponding visual attention features, and then uses joint representations to calculate gates which decide which low-level concept and visual feature can pass to higher levels. By applying the fuse-select procedure, our proposed model with GHA outperforms other image captioning models that employ CNNs. 


\begin{figure}[t]
\centering
\includegraphics[width=\textwidth]{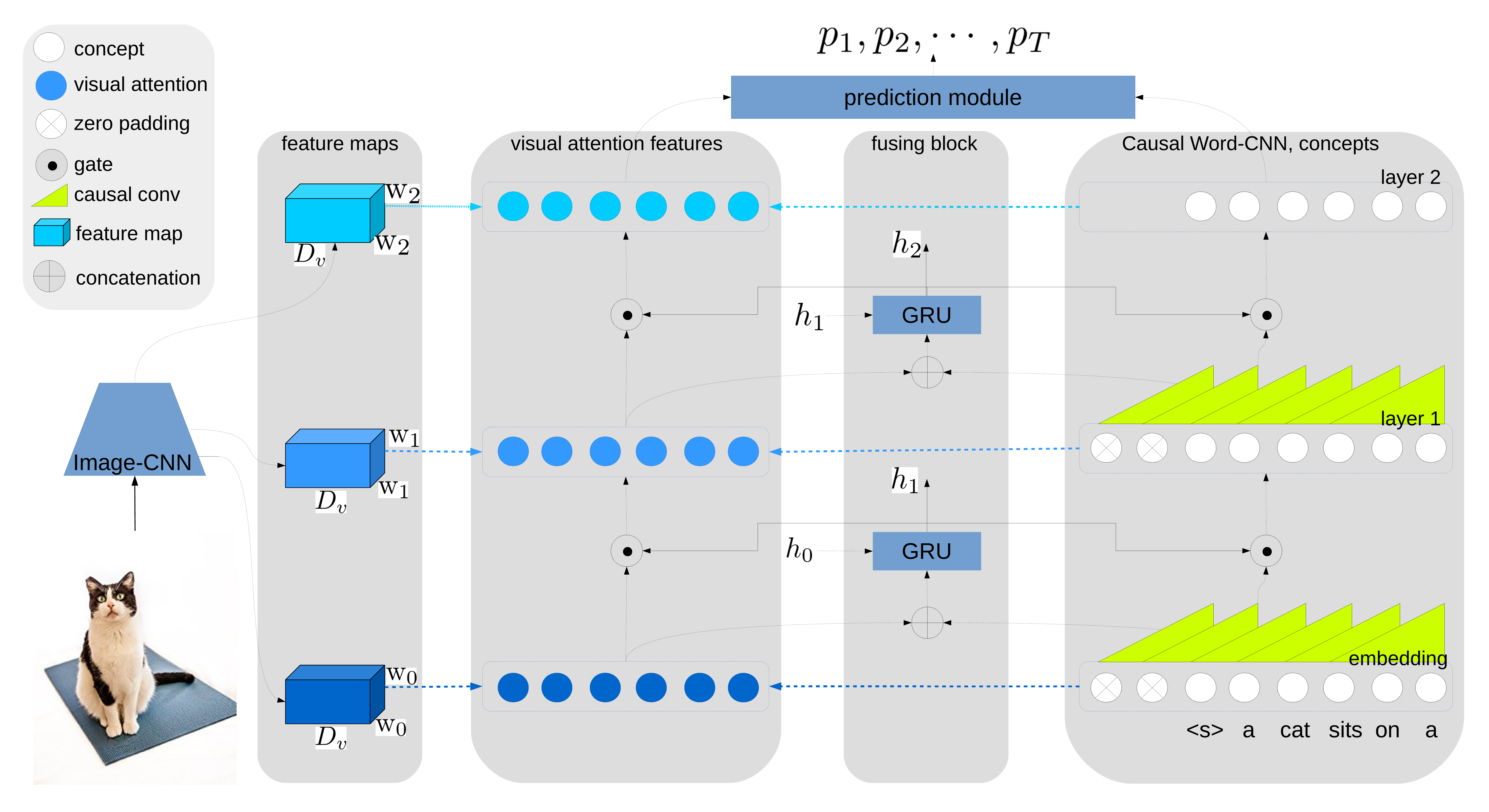}
\caption{An overview of our GHA. Here we show our captioning model with 2 causal convolutional layers. The dash arrows represent calculating the visual attention features using Eq.~\ref{eq1}. The GHA can use different convolutional feature maps, which are represented by different colors. The fusing block selects the relevant visual and concept information to pass to the higher layer.
}\label{GHAfig}
\end{figure}

\section{Gated Hierarchical Attention}
In this section, we present our captioning model with gated-hierarchical attention (GHA).
There are four parts in our proposed GHA model: 1) image encoder, 2) visual attention, 3) causal convolutional decoder, and 4) fusing block. 
An overview of the proposed GHA model is shown in Figure \ref{GHAfig}. 
The image encoder and word decoder drive the visual attention to extract relevant features from the image feature maps.
At each level, the fusing block plays the role of information fusion and selection, which allows interactions between Image-CNNs and Word-CNNs. 
There are gates between layer $l-1$ and layer $l$, which controls the bottom-up flow of visual and language information. The gates depend on both low-level visual and language features, and more details are in the following sections.


\subsubsection{Image encoder.}
The goal of the image encoder is to obtain a feature representation of an image. 
In this paper, we adopt an Image-CNN to extract visual features, which is similar to other image captioning models \cite{spatt,NIC,languagecnn,convimagecap}. 
The main difference is that our GHA is able to use convolutional feature maps at different levels, which is shown in Figure \ref{GHAfig}. Different convolutional feature maps have different receptive fields, and a feature map at lower layer can ``see'' smaller objects and more details of an image, while high-layer feature maps ``see'' parts of large objects or even the entire object. Therefore, applying multiple feature maps is able to benefit object recognition and word prediction \cite{learn2payatt}.
The image feature maps at different levels are projected to the same channel-dimension using a learnable linear transformation.
Thus, an image is represented by a set of feature maps $\mathbf{v} = \{v^l_{ij}|i=1,\ldots,w;j=1,\ldots,h;l=0,\ldots, L\}$, where $v^l_{ij}\in \mathbb{R}^{D_v}$ is the image feature vector at image position $(i,j)$ for layer $l$.

\subsubsection{Causal Word-CNNs.}
The goal of Word-CNNs is to merge words to obtain a high-level representation of the sentence. 
Since we do not know the future words during inference time, we employ causal convolution, which only considers the previous words. In this paper, the Word-CNNs stack multiple causal convolutional layers (see Figure \ref{GHAfig}, which shows a decoder with kernel size 3). 
The Word-CNN produces a set of concept representations at different levels, $\mathbf{c}=\{c^l_t|t=1,\ldots, T;l=0,\ldots, L\}$, where $c^l_i\in \mathbb{R}^{D_c}$ is the concept vector for level $l$ and position $i$. $L$ is the number of the convolutional layers in our decoder CNN, and $l=0$ is the word embedding layer. $T$ is the length of a caption.

For the activation function, gated linear units (GLU) show more advantages than ReLU in NLP \cite{glulanguage}, and other convolutional image captioning models also employ GLU \cite{convimagecap,cnnpluscnn}. 
In a standard GLU, both the gates and linear activations are computed using the hidden states of the previous layer of the decoder. 
For our GHA, we use a variant of GLU where the previous decoder layer is used to drive the linear unit and the fusing block is used to drive the gate,
\begin{equation}
\begin{aligned}
c^l_{1:T} &= \underbrace{(\mathbf{W}_a\ast c^{l-1}_{1:T})}_{\text{linear unit}} \odot \underbrace{\textbf{sigmoid}\left( \mathbf{W}_b\ast h^{l}_{1:T}\right)}_{\text{gate}},
\end{aligned}
\end{equation}
where $c^l_{1:T}$ is the concepts at the $l$-th layer of the decoder, and
 $h^l_{1:T}$ is the hidden state of the fusing block (described later).
$\ast$ denotes convolution, $\odot$ is element-wise multiplication, and $\{\mathbf{W}_a,\mathbf{W}_b\}$ are trainable weights. The advantage of using $h^l_{1:T}$ instead of $c^{l-1}_{1:T}$ for the gate is that the visual attention feature $\hat{v}^{l}_{t}$ are able to decide which concept features can pass to higher levels, resulting in more significant interaction between concepts and visual features.

\subsubsection{Visual attention.}
The visual attention module 
aims to learn the correspondence between concepts and image areas. In our  GHA, the visual attention feature $\hat{v}_{t}^{l}$ at $l$-th level is composed of two parts\footnote{For the word embedding layer $l=0$, 
only the first part is used.
}: 1) the current level visual attention feature $\tilde{v}_{t}^{l}$, 2) the previous level attention feature $\hat{v}^{l-1}_{t}$. The current-level attention feature $\tilde{v}^{l}_{t}$ is calculated 
using the concepts $c_t^l$ at the same level:
\begin{equation}\label{eq1}
\begin{aligned}
s_{ij}^{lt} = \tfrac{1}{\sqrt{D_c}}(v_{ij}^{l})^\text{T}\mathbf{W}c_{t}^l,
\quad 
a_{ij}^{lt} = \frac{e^{s_{ij}^{lt}}}{\sum_{i=1}^{w}\sum_{j=1}^{h}e^{s_{ij}^{lt}}}, 
\quad
\tilde{v}^{l}_{t} = \sum_{i=1}^w\sum_{j=1}^h a_{ij}^{lt}\cdot v^l_{ij},
\end{aligned}
\end{equation}
where $\mathbf{W}\in \mathbb{R}^{D_v\times D_c}$ is a parameter matrix. Note that, similar to \cite{allatt}, we apply a scaling factor $\frac{1}{\sqrt{D_c}}$ to $s_{ij}^{lt}$ since it might take large values.

The final visual attention feature $\hat{v}^{l}_{t}$ is computed as follows:
\begin{equation}
\hat{v}^{l}_{t} = \tilde{v}^{l}_{t} + \textbf{sigmoid}\left(\mathbf{W}_vh_t^l\right)\odot \hat{v}^{l-1}_{t},
\end{equation}
where $h_t^l\in \mathbb{R}^{D_h}$ is the hidden state of the fusing block, and $\mathbf{W}_v\in \mathbb{R}^{D_v\times D_h}$ are the trainable parameters. By using gates, GHA is able to select lower-level visual attention features ($\hat{v}^{l-1}_t$) and fuse them with current (higher) level features ($\tilde{v}^l_t$).

\subsubsection{Fusing block.} 
The motivation of our GHA is to generate different level visual attention features for different concepts and pass low-level features to higher levels. To achieve the goal, we use a GRU to fuse the low-level concepts and the corresponding visual attention features. 
Concatenating the visual attention features and concepts $x_t^l=[\hat{v}^{l-1}_{t},c_t^{l-1}]$, 
the hidden state of the fusing block $h_t^l$ is computed as follows:
\begin{equation}\label{eq4}
\begin{aligned}
r_t^l &= \textbf{sigmoid}\left(\mathbf{W}_r[h_t^{l-1},x_t^l]\right),
& {\tilde{h}}_t^l &= \textbf{tanh}\left(\mathbf{W}_{\tilde{h}} \cdot[r_t^l\odot h_t^{l-1},x_t^l]\right), \\
z_t^l &= \textbf{sigmoid}\left(\mathbf{W}_z[h_t^{l-1},x_t^l]\right), &
h_t^l &= (1-z_t^l)\odot h_t^{l-1}+z_t^l \odot \tilde{h}_t^{l-1},
\end{aligned}
\end{equation}
where 
the initial hidden state is $h_t^0 = \mathbf{0}$. Note that we apply GRUs in a different way. Normally, GRUs are employed to modeling sequences, and thus the recurrent path is along with the time axis. In contrast, in our fusing block the recurrent path is along the hierarchical levels, from bottom to top. 

\subsubsection{Prediction and Loss Function.}
We employ a 3-layer MLP to predict the next word at sentence position 
$t$. Dropout layers 
are also applied to mitigate overfitting. At the $t$-th position, the prediction MLP takes $\hat{v}^{L}_{t}$ and $c_{t}^L$ as input, and outputs the probability distribution $p_{t}$ of the next word. The loss function is the  cross-entropy loss between $p_t$ and the ground-truth words in the sentence.

\section{Experiments}\label{exp}

\subsection{Dataset and Preprocessing}

In this paper, we conduct experiments on MSCOCO dataset \cite{mscocodataset}, which contains 82,783 training and 40,504 validation images, where each image has at least 5 captions. We use the same train/validation/test split as \cite{deepVS} (denoted as ``Karpathy split''), which uses 5,000 images for validation, 5,000 images for testing and the remaining images for training. 

We drop the words that occur less than 6 times and obtain a vocabulary with 9,489 words, including 3 special tokens: starting token $<$start$>$, ending token $<$end$>$ and unknown token $<$unk$>$. During training time, the images are resized to 256$\times$256 and then we randomly crop a patch with the size of 224$\times$224. Data is augmented using random horizontal flipping. During testing time, we resize all testing images to 224$\times$224 and directly feed them into our trained model with the starting token. For the model that we submit to the testing server, we use our train$+$val split to train it.

\subsection{Implementation Details}

\subsubsection{Baseline Model.} To compare the effect of using GHA, we test baseline models where GHA is removed, and only visual attention at the top level is used, similar to \cite{cnnpluscnn}.
The decoder in our baseline model has $L=6$ convolutional layers, and the kernel size of each convolutional layer is 3 (this baseline is denoted as ``Base-6-3''). Therefore each unit at the top layer can ``see'' 13 words, which is larger than the average length (11.6) of the captions in MSCOCO. The image encoder is Resnet101 without fully connected layers, therefore, $w=h=7$ and $D_v=2048$.  The prediction module has 3 fully connected layers, and each hidden layer has 4096 units with ELU activation function \cite{elu}. After each hidden layer we employ a dropout \cite{dropout} layer with the keep probability of 0.5. We use $D_c=300$.

\subsubsection{GHA Model.}
For the models with GHA, the number \wqz{of} hidden units of the GRU is 512, and we apply the same feature map---the output of conv5\_x in Resnet101 to all levels. The same number of layers and kernel size is used for the decoder, and our GHA model is denoted as ``GHA-6-3''.
For multi-scale GHA (MS-GHA), we use 3 different feature maps computed by Resnet101: the outputs of conv3\_x, conv4\_x and conv5\_x, which have sizes of 28$\times$28$\times$512, 14$\times$14$\times$1024 and 7$\times$7$\times$2048. For easy implementation and reduce the computation complexity, we first transfer $\mathbb{R}^{512}$ and $\mathbb{R}^{1024}$ into $\mathbb{R}^{2048}$ and apply average pooling with $k=2$ to the 28$\times$28 feature map. After this, the feature maps become $14\times14\times2048$, $14\times14\times2048$ and $7\times7\times2048$. Our MS-GHA models also have $L=6$ convolutional layers plus an word embedding layer. For the word embedding layer and the first decoder level we use the conv3\_x feature map to compute the attention features. For the subsequent 3 decoder levels, the conv4\_x feature map are used for attention, and the last 2 decoder levels 
use the conv5\_x feature map. 
The models with GHA employ the same prediction module as the baseline model.

\subsubsection{Training and Inference.}
We apply the Adam optimizer \cite{adamop} to train all models and use the pre-trained Resnet101 on the ImageNet dataset. The learning rate is fixed during training to $1\times10^{-5}$ for the Resnet101 parameters, and $3\times10^{-4}$ for the other parameters.  
We train all models for 50 epochs and report the performance of the best model during training.
During inference, we apply beam search \cite{convimagecap} 
with a beam width of 3. The maximum length of the generated sentences is 20 for all models.


\begin{table}[t]
\caption{Performance on MSCOCO Karpathy test split. $\star$ denotes applying attributes or semantics, $\dag$ denotes using reinforcement learning and $\ddag$ means ensembling multiple trained models. B, M, R, C, S respectively represent BLEU \cite{bleu}, METEOR \cite{M}, ROUGE \cite{R}, CIDEr \cite{C} and SPICE \cite{spice} metrics.}\label{tab2}
\centering
\scalebox{1.0}{
\begin{tabular}{c|c|c|c|c|c|c|c|c|c}
\hline
\multicolumn{2}{c|}{\textbf{Method}} &B-1 &B-2 &B-3 &B-4 &M &R &C &S \\
\hline
\multirow{9}{*}{LSTM} &Hard-Att \cite{spatt} &0.718 &0.504 &0.357 &0.250 &0.230 &- &- &- \\
&ATT-FCN$^\star\ddag$ \cite{semanticatt} &0.709 &0.537 &0.402 &0.304 &0.243 &- &- &- \\
&SCN$^{\star\ddag}$ \cite{SCN} &0.741 &0.578 &0.444 &0.341 &0.261 &- &1.041 &- \\
&Adaptive$^{\ddag}$ \cite{when2look} &0.742 &0.580 &0.439 &0.332 &0.266 &- &1.085 &- \\
&SCST:Att2in \cite{scst} &- &- &- &0.313 &0.260 &0.543 &1.013 &- \\
&SCST:Att2in$^{\dag}$ \cite{scst} &- &- &- &0.333 &0.263 &0.553 &1.114 &- \\
&MSM$^{\star\ddag}$ \cite{attributesboosting} &0.734 &0.567 &0.430 &0.326 &0.254 &0.540 &1.002 &0.186 \\
&PG-BCMR$^{\dag}$ \cite{bmcr} &0.754 &0.591 &0.445 &0.332 &0.257 &0.550 &1.013 &- \\
&AG-CVAE \cite{cvae} &0.732 &0.559 &0.417 &0.311 &0.245 &0.528 &1.001 &0.179\\
\hline
\multirow{2}{*}{CNN} &CNN+RHN \cite{languagecnn} &0.723 &0.553 &0.413 &0.306 &0.252 &- &0.989 &0.183 \\
&CNN$+$Att \cite{convimagecap} &0.722 &0.553 &0.418 &0.316 &0.250 &0.531 &0.952 &0.179 \\
&CNN+CNN \cite{cnnpluscnn} &0.685 &0.511 &0.369 &0.267 &0.234 &0.510 &0.844 &- \\
\hline
\multirow{2}{*}{Ours} &Base-6-3 &0.702 &0.531 &0.396 &0.295 &0.246 &0.520 &0.923 &0.174 \\
& GHA-6-3 &0.733 &0.564 &0.426 &0.321 &0.255 &0.538 &0.999 & \bf{0.189} \\
\hline
\end{tabular}
}
\end{table}

\begin{table}[t]
\caption{Performance on MSCOCO test dataset. The performance of other methods is from the latest version of their papers and the human performance is from the leaderboard (\url{
http://mscoco.org/dataset/\#captions-leaderboard}). The metrics are the same as Table \ref{tab2}. \textbf{c5} denotes that one image has 5 reference captions and \textbf{c40} represents one image corresponds to 40 reference captions. Note that the test dataset contains 40,775 images which is different from the Karpathy test split (5,000 images).}\label{tab3}

\centering
\scalebox{0.8}{
\begin{tabular}{c|ccccccc|ccccccc}
\hline
\multirow{2}{*}{\textbf{Method}} &\multicolumn{7}{c|}{\textbf{c5}} &\multicolumn{7}{c}{\textbf{c40}} \\
\cline{2-15}
&B-1 &B-2 &B-3 &B-4 &M &R &C &B-1 &B-2 &B-3 &B-4 &M &R &C \\
\hline
Human &0.663 &0.469 &0.321 &0.217 &0.252 &0.484 &0.854 &0.880 &0.744 &0.603 &0.471 &0.335 &0.626 &0.910 \\
ATT-FCN \cite{semanticatt} &0.731 &0.565 &0.424 &0.316 &0.250 &0.535 &0.943 &0.900 &0.815 &0.709 &0.599 &0.335 &0.682 &0.958 \\
Review Net \cite{reviewnet} &- &- &- &- &- &- &- &- &- &- &0.597 &0.347 &0.686 &0.969 \\
\hline
MSM \cite{attributesboosting} &0.787 &0.627 &0.476 &0.356 &0.270 &0.564 &1.16 &0.937 &0.867 &0.765 &0.652 &0.354 &0.705 &1.18 \\
SCN \cite{SCN} &0.740 &0.575 &0.436 &0.331 &0.257 &0.543 &1.003 &0.917 &0.839 &0.739 &0.631 &0.348 &0.696 &1.013 \\
SCST:Att2in \cite{scst} &- &- &- &0.344 &0.268 &0.559 &1.112 &-&- &-&-&-&-&- \\
CNN+Att \cite{convimagecap} &0.715 &0.545 &0.408 &0.304 &0.246 &0.525 &0.910 &0.896 &0.805 &0.694 &0.582 &0.333 &0.673 &0.914 \\
GHA-6-3 (ours) &0.729 &0.560 &0.419 &0.313 &0.252 &0.533 &0.954 &0.937 &0.818 &0.708 &0.598 &0.341 &0.683 &0.963 \\
\hline
\end{tabular}
}
\end{table}

\subsection{Results on Karpathy test split}

To compare with other methods, we first categorize the methods into 2 categories: 1) LSTM-based methods, and 2) CNN-based methods. Our proposed model is CNN-based.   Results on the Karpathy test split are shown in Table \ref{tab2}. 

Our GHA significantly improve the performance over the baseline model --  BLEU-1,2,3,4 (B-1, etc) scores increase by around 0.03 (relative improvement of 4.4\%, 6.2\%, 7.6\%, and 8.8\%), the CIDEr score increases from 0.923 to 0.999  (increase of 8.2\%), and the SPICE score increases by 0.015 (increase of 8.6\%), which are all significant improvements. 
%
Our GHA model outperforms other CNN-based models (CNN+RHN, CNN+Att and CNN+CNN) on all metrics, which suggests that the usefulness of our proposed GHA. Note that the goals of our convolutional decoder and the language CNNs in \cite{languagecnn} are different. In \cite{languagecnn}, language CNNs are employed to make the LSTMs ``see'' more words, while our convolutional decoder is used to mimic the hierarchical structures of sentences (and therefore does not use LSTMs). Although CNN+Att and CNN+CNN apply attention at each decoder level, their simple approach without gates cannot filter out useless information, leading to worse performance.

Compared with Hard-Att \cite{spatt}, which applies an MLP to compute the attention weights, our baseline model obtains better results except for the B-1 score. By introducing our proposed GHA, the model obtains comparable performance on all metrics with LSTM-based models MSM \cite{attributesboosting} and AG-CVAE \cite{cvae}, which employ attributes or semantics. Note that applying attributes or semantics requires an extra branch to first predict the attributes or semantics, and the generated captions are highly related to the predicted semantics. If the semantics are incorrect or noisy, the captions could contain errors.
SPICE correlates well with human judgements \cite{LEIC}, and our GHA obtains higher SPICE than other methods.

Our GHA performs a little worse than SCST \cite{scst} and PG-BCMR \cite{bmcr}, which  use reinforcement learning (RL) to directly maximize the evaluation metrics.  The GHA scores are lower by 0.01-0.02 than the models that apply RL.  However, note that most of this improvement is from the RL method, because it is employed to fine-tune a model trained with cross-entropy loss.  RL is able to suppress the words that cannot improve the evaluation metrics, and encourage the words that increase the metric scores, thus yielding better performance on these metrics. 
However, applying RL requires sampling during training, which is time-consuming.


\subsection{Results on online test set}

Table \ref{tab3} shows the comparison of the captioning methods on the online test dataset. Looking at CIDEr scores, our GHA model outperforms CNN+Att \cite{convimagecap} and ATT-FCN \cite{semanticatt}, but obtain lower scores than MSM \cite{attributesboosting}, SCN \cite{SCN} and SCST:Att2in \cite{scst}. 
%
The results are mostly consistent with the Karpathy test split, although the CNN-based models perform slightly worse on the testing server.
This is possibly due to overfitting since the CNN-based models have 
more trainable parameters compared to LSTM-based methods.
For example, CNN+Att \cite{convimagecap} has 20M parameters, while DeepVS \cite{deepVS} only has 13M parameters.  Therefore to some extent, CNN-based models are more likely to overfit the dataset.

The BLEU, METEOR, ROUGE, and CIDEr (BMRC) metrics often do not correlate to human judgments, because they just consider the words or n-grams in the reference captions \cite{LEIC}. Under these metrics, the human captions actually have worse performance than most models. However, looking at the generated captions reveals that most models just memorize the common words and phrases to describe images, and thus some similar images will have the same caption. In contrast, human annotations tend to have varying captions, since the background knowledge of each person varies, leading to lower BMRC metrics. SPICE \cite{spice} has been shown to be better correlated with human judgement \cite{LEIC}. Unfortunately, the SPICE metric is currently not available from the online test server.

\begin{figure}[t]
\centering
\includegraphics[width=\textwidth]{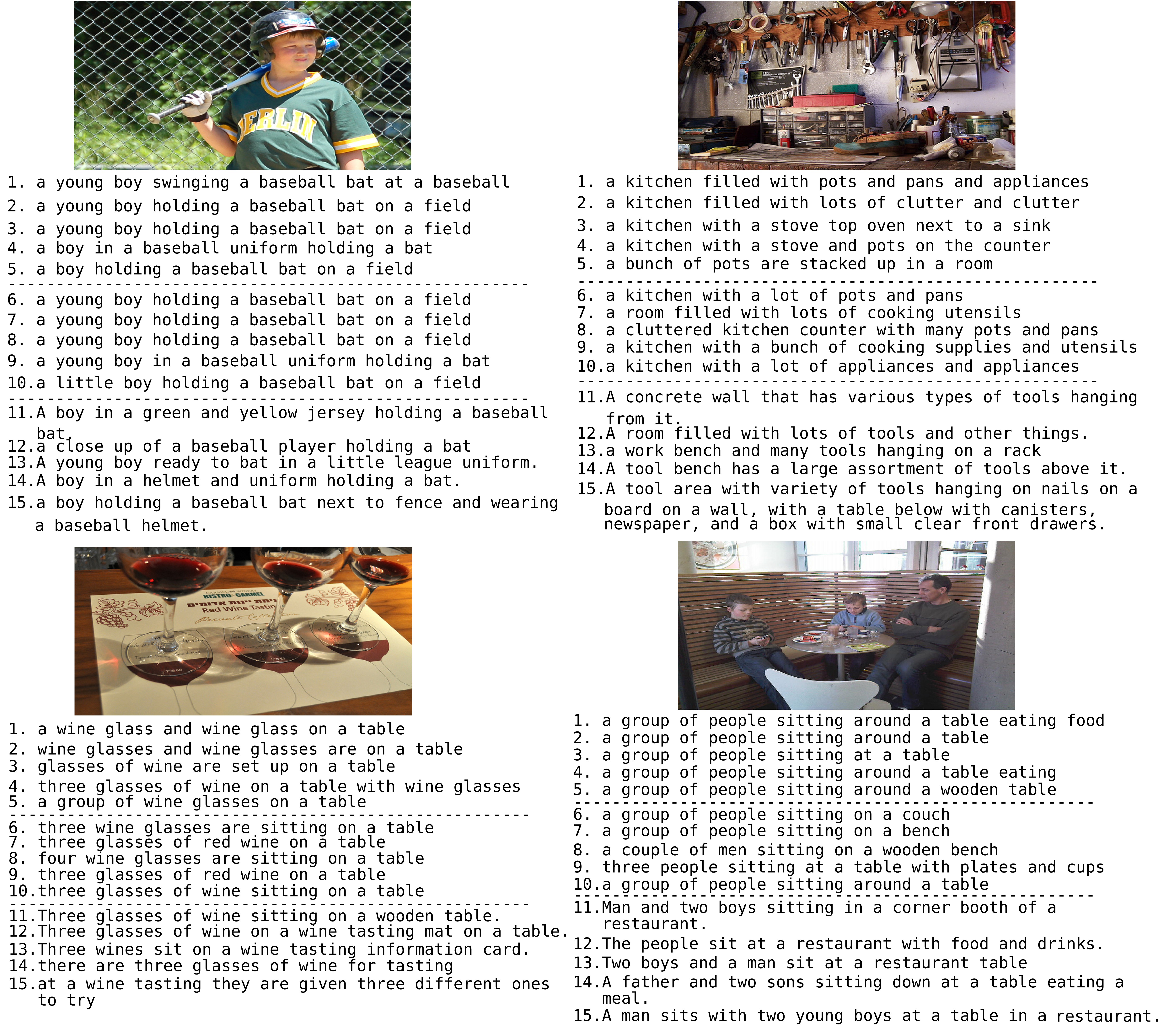}
\vspace{-2.5em}
\caption{Examples of the generated captions by our GHA model and base model. The baseline models are: (1) Base-6-3, (2) Base-6-5, (3) Base-6-7, (4) Base-10-3, and (5) Base-6B-3.  The GHA models are: (6) GHA-6-3, (7) GHA-6-5, (8) GHA-6-7, (9) GHA-10-3, and (10) GHA-6B-3. The human (ground-truth) annotations are: (11)-(15).}\label{examples}
\end{figure}

\begin{figure}[t]
\centering
\includegraphics[width=\textwidth]{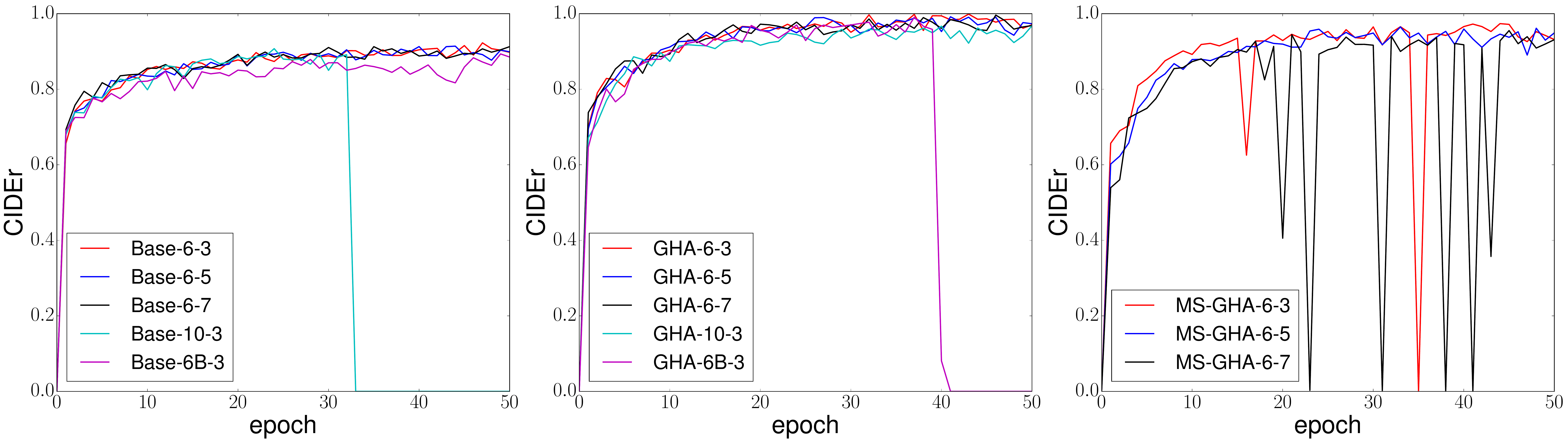}
\caption{CIDEr performance during training. Left: baseline models (without the proposed GHA). Middle: models with GHA. Right: models with multi-scale GHA.
}\label{collapse}
\end{figure}

\begin{figure}[t]
\centering
\includegraphics[width=\textwidth, height=0.55\textheight]{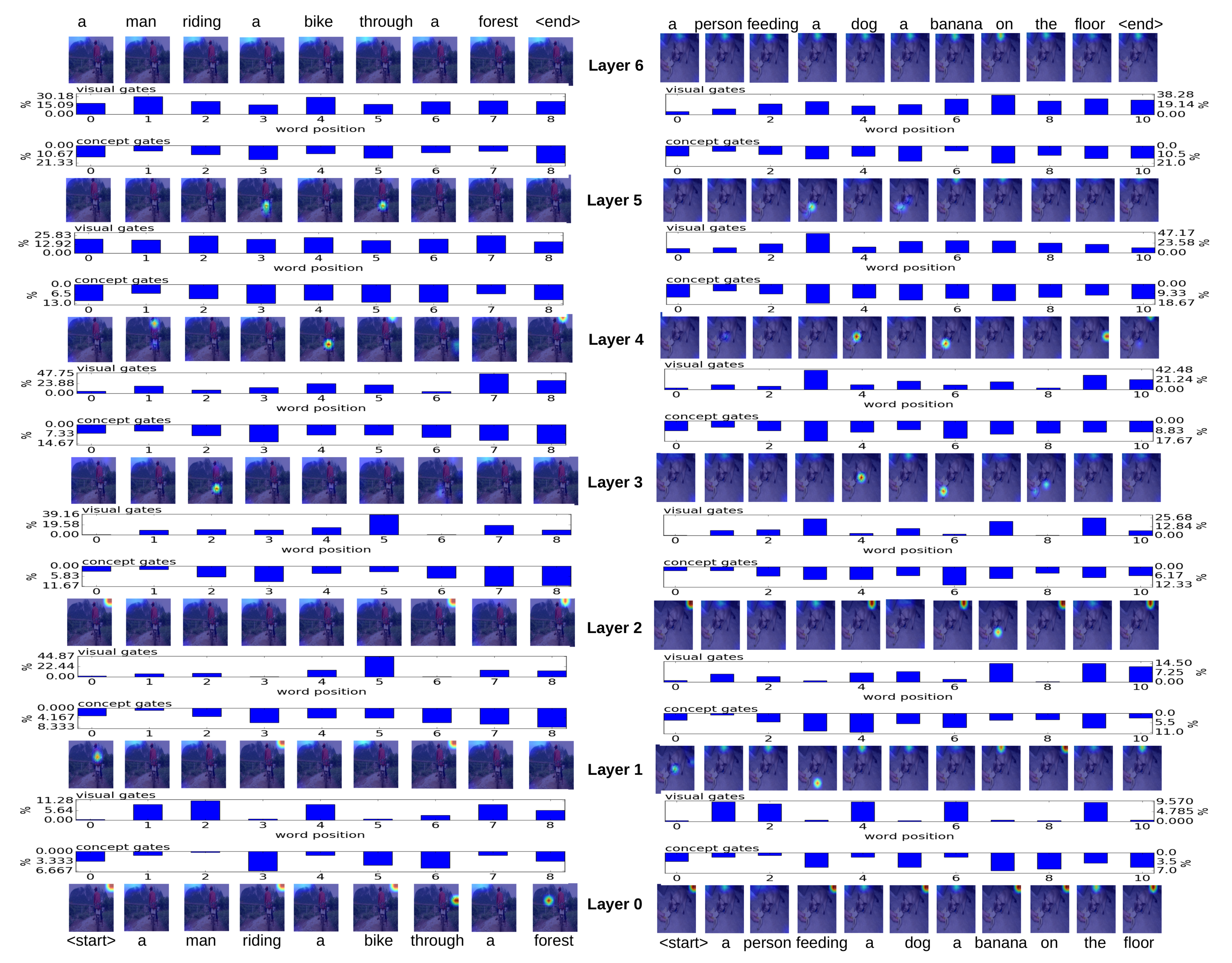}
\caption{Visualizing the attention weights and gate states of our GHA.   The attention weights are visualized as a heat map on the image.  The bar charts show the percentage of visual and concept gates that are ``on'', allowing lower-level information through.  The bottom text are the input words and the top text are the predicted words.}\label{att2}
\end{figure}

\comments{
\begin{figure}[t]
\centering
\includegraphics[width=0.8\textwidth]{figures/failure.pdf}
\caption{Failure cases. In some failure cases, the model generates incorrect descriptions (the first row). In other cases, although the descriptions are correct, the attention maps are not reasonable (the fourth row).}\label{failure}
\end{figure}
}

\begin{table}[t]
\caption{Model variants used in the ablation study. ``M-$l$-$k$'' represents a captioning model M, where $l$ represents the number of convolutional layers of the decoder and $k$ is the kernel size of each layer. B represents a model that applies bottleneck connections. ``Base'' represents a model without GHA, which only computes visual attention features at the top layer. GHA means using our proposed GHA and MS-GHA denotes multi-scale GHA, which applies different convolutional feature maps of the image encoder. In the table, $[k, D_c]$ represents the structure of the filter in the decoder, where $D_c$ is the dimension of the concept space.} \label{tab1}

\centering
\scalebox{0.9}{
\begin{tabular}{c|c|c|c|c|c|c}
\hline
Base-6-3 & Base-6-5 & Base-6-7 & Base-10-3 & Base-6B-3 & GHA-6-3 & GHA-6-5  \\
\hline
[3, 300]$\times$6 &[5, 300]$\times$6 &[7, 300]$\times$6 &[3, 300]$\times$10 &$\left[\begin{array}{cc}1,&300\\ 3, &300\\ 1, &300\end{array}\right]\times$6 &[3, 300]$\times$6 &[5, 300]$\times$6 \\

\hline
\hline
GHA-6-7 & GHA-10-3 & GHA-6B-3 & MS-GHA-6-3 & MS-GHA-6-5 & MS-GHA-6-7 &- \\
\hline
[7, 300]$\times$6 &[3, 300]$\times$10 &$\left[\begin{array}{cc}1,&300\\ 3, &300\\ 1, &300\end{array}\right]\times$6 &[3, 300]$\times$6 &[5, 300]$\times$6 &[7, 300]$\times$6 &- \\
\hline
\end{tabular}
}
\end{table}

\begin{table}[t]
\caption{Performance of models with different kernel sizes and number of layers on Karpathy test split. The metrics are the same as Table \ref{tab2}.}\label{tab4}
\centering
\scalebox{1.0}{
\begin{tabular}{c|c|c|c|c|c|c|c|c|c}
\hline
\multicolumn{2}{c|}{\textbf{Method}} &B-1 &B-2 &B-3 &B-4 &M &R &C &S \\
\hline
\multirow{5}{*}{baseline}
& Base-6-3 &0.702 &0.531 &0.396 &0.295 &0.246 &0.520 &0.923 &0.174 \\
& Base-6-5 &0.703 &0.532 &0.395 &0.294 &0.243 &0.518 &0.914 &0.172 \\
& Base-6-7 &0.708 &0.536 &0.397 &0.293 &0.241 &0.518 &0.913 &0.172 \\
& Base-10-3 &0.703 &0.534 &0.397 &0.296 &0.241 &0.517 &0.908 &0.171 \\
& Base-6B-3 &0.701 &0.530 &0.392 &0.288 &0.239 &0.515 &0.894 &0.172 \\
\hline

\multirow{5}{*}{GHA}
& GHA-6-3 &0.733 &0.564 &0.426 &0.321 &0.255 &0.538 &0.999 &0.189 \\
& GHA-6-5 &0.732 &0.561 &0.422 &0.316 &0.254 &0.534 &0.992 &0.189 \\
& GHA-6-7 &0.726 &0.559 &0.421 &0.318 &0.255 &0.535 &0.996 &0.185 \\
& GHA-10-3 &0.720 &0.550 &0.412 &0.308 &0.252 &0.530 &0.968 &0.184 \\
& GHA-6B-3 &0.725 &0.559 &0.423 &0.320 &0.255 &0.535 &0.991 &0.185 \\
\hline 

\multirow{3}{*}{MS-GHA}
& MS-GHA-6-3 &0.717 &0.549 &0.413 &0.310 &0.252 &0.529 &0.974 &0.185 \\
& MS-GHA-6-5 &0.724 &0.552 &0.412 &0.309 &0.250 &0.529 &0.965 &0.182 \\
& MS-GHA-6-7 &0.718 &0.547 &0.407 &0.303 &0.250 &0.527 &0.955 &0.186 \\
\hline
\end{tabular}
}
\end{table}

\begin{table}[h]
\caption{The performance of GHA with different feature maps of Resnet101 on Karpathy test split.}\label{tab5}
\centering
\scalebox{1.0}{
\begin{tabular}{c|c|c|c|c|c|c|c|c}
\hline
Method &Convolutional map &B1 &B2 &B3 &B4 &M &R &C \\
\hline
\multirow{3}{*}{GHA-6-3} 
&conv3\_x &0.600 &0.416 &0.291 &0.208 &0.191 &0.446 &0.597 \\
&conv4\_x &0.709 &0.537 &0.399 &0.298 &0.245 &0.512 &0.921 \\
&conv5\_x &0.733 &0.564 &0.426 &0.321 &0.255 &0.538 &0.999 \\
\hline
\end{tabular}
}
\end{table}

\subsection{Ablation Study and Analysis}

To explore the impact of the depth and kernel size on our proposed models, we run an ablation study using  model variants with different layers and kernel sizes, as listed in Table \ref{tab1}. 
For the model with 6 bottlenecks, which has an 18-layer decoder with 6 shortcut connections, we only use kernel size 3, because the best performances of the GHA and baseline models 
are obtained by using kernel size 3. The performance of all models is shown in Table \ref{tab4}.

\subsubsection{Effectiveness of GHA.}
The models employing our 
GHA performs much better than the baseline models that do not use GHA. Applying GHA increases the BLEU scores by around 0.03 (a relative improvement of 4.4\%, 6.2\%, 7.6\%, and 8.8\%), and the CIDEr score by more than 0.07 (increase of 8.2\%), and the SPICE score by 0.015 (increase of 8.6\%), which are all significant improvements. 
However, the MS-GHA models performs worse than single-scale GHA models. One possible reason is that our image encoder is Resnet101, which contains shortcut connections that allow the top convolutional feature map to be fused with the low-level features. Therefore, the shortcut connections in residual nets work in the same manner to our MS-GHA model. Thus, it is unnecessary to employ multi-scale attention. Another possible reason is that MS-GHA introduces low-level visual features, such as edges and corners, which do not contain any semantic information. \wqz{Table \ref{tab5} shows the results of applying different feature maps of Resnet101. 
Applying low-level features could result in worse performance. In particular using conv3\_x, the BMRC scores decrease drastically.}

Some examples of generated captions are shown in Figure \ref{examples}. For the simple scenario with a few objects, all the models are able to recognize the objects, their relationships and the actions. However, if there are many different objects in an image, the models sometimes provide incorrect descriptions. Moreover, object counting can be difficult for the models. Introducing more visual information could benefit counting objects.  For example, the models with GHA are able to recognize ``three glasses'' and ``three people'', while the baseline models without GHA are likely to ignore the numbers, even though the training data provide the numbers in the captions.

\subsubsection{Is deeper better?} In many cases, in particular object recognition, deeper neural networks perform better than shallow ones. Our experiments show that for image captioning, deeper convolutional decoders are not necessary. When the decoder becomes deeper, the performance becomes slightly worse. The goal of stacking more convolutional layers in the decoder is to increase the receptive field to ``see'' more words, however the average length of captions in MSCOCO is 11.6 words and our 6-layer decoder with kernel size 3 is able to ``see'' 13 words, which is enough to model the captions. Interestingly, if we fix the number of layers and increase the kernel size (e.g., Base-6-3 to Base-6-5 to Base-6-7), the performance marginally decreases (CIDEr drops from 0.923 to 0.914 and 0.913).

Another drawback of deeper decoders is that they are not stable during training. We found that the models with deeper decoders could collapse during training.  In some epochs, the model predictions are all the same 
even though the input images are different, causing the CIDEr score to drop drastically  (e.g., Figure \ref{collapse}). The models that employ MS-GHA have the same problem, but finally they will stabilize and generate satisfying captions. One possible reason for this problem is that it is difficult to train deeper models. Moreover, deeper convolutional decoders require more training and testing time.

\subsubsection{Attention and Gate Visualization.} 
We visualize the attention maps and gates at each decoder layer (see Figure \ref{att2}). To visualize the $7\times7$ attention map, we first upsample it to $224 \times 224$, and then overlay it on the original image. We also show the percentage of the visual and concept gates that are ``on'', allowing information to flow from lower levels to higher levels (see the bar charts in Figure \ref{att2}).  The gate values range from 0 (completely off) to 1 (completely on). The gate is considered on if its value is above 0.65 and 0.25 for the visual and concept gates, respectively.  Note that there are 2,048 visual gates and 300 concept gates at each decoder level (one for each channel) and some gates could be on while others could be off at the same time. 

From the attention maps, we can find that higher decoder levels are more likely to ``see'' the entire image and more gates are on at higher decoder levels.  The gates are able to filter out the visual information when the model pays attention to the non-object areas, such as corners of an image. \Eg, in Figure \ref{att2} (left), the first, fourth and sixth attention maps at layer 0 pay attention to the non-object areas, and the corresponding visual gates are off to filter out the information. Interestingly, if the attention weights are flat, the visual gates are likely to be on -- thus the global representation of an image is able to benefit word prediction. For the correct object areas, some visual gates are on, which is reasonable. In Figure \ref{att2} (left), the second and fifth attention maps at layer 4 pay attention to the person and the bike in the image, respectively, and the visual gates turn on, resulting in predicting the words ``man'' and ``bike''. This also occurs Figure \ref{att2} (right) -- in the fifth and seventh attention maps, dog and banana are attended when some visual gates are on. 

Looking at the trends of the bar charts, our GHA exhibits a similar behavior as \cite{when2look}: if more visual gates are off, then more concept gates will be on, in particular at lower decoder layers. Moreover, to predict a word that corresponds to an object in the image, such as ``man'' and ``bike'', more visual gates are open. In contrast, to predict the words ``a'', ``through'' and the ending token ``$<$end$>$'', more concept gates are on, which indicates that those words are more likely to depend on context instead of image features. Note that \cite{when2look} switches between concept and visual features only at the highest feature-level, whereas our GHA performs the switching hierarchically, throughout the bottom-up flow of information.


\section{Conclusion and Future Work}
\vspace{-0.5em}
In this paper, we have presented  gated hierarchal attention (GHA), which significantly improves the performance of the convolutional captioning models. Applying GHA, we obtained comparable performance with the state-of-the-art models, such as reinforcement learning methods, attributes boosting models, and LSTMs. We also analyzed the impact of hyper-parameters, such as depth and kernel size, and our experiments suggest that it is not necessary to use a deeper convolutional decoder. 

Although we have conducted extensive experiments to show the abilities of GHA and convolutional decoders for image captioning, some issues should be solved in the future. One is that current attention mechanisms have limited ability to reveal the relationship between objects, therefore the image captioning models could generate strange descriptions. Hence, one direction of the future work could be learning better relationship representations to improve the quality of the generated captions. Another issue is that current captioning models could generate the same description for different images by using common words, which lacks distinctiveness. Therefore, another direction of the future work could be generating distinctive descriptions.

\bibliographystyle{splncs04}
\bibliography{myref}

\end{document}